# Robust Lane Detection through Self Pre-training with Masked Sequential Autoencoders and Fine-tuning with Customized PolyLoss

Ruohan Li#, Yongqi Dong#

*Abstract*—Lane detection is crucial for vehicle localization which makes it the foundation for automated driving and many intelligent and advanced driving assistant systems. Available vision-based lane detection methods do not make full use of the valuable features and aggregate contextual information, especially the interrelationships between lane lines and other regions of the images in continuous frames. To fill this research gap and upgrade lane detection performance, this paper proposes a pipeline consisting of self pre-training with masked sequential autoencoders and fine-tuning with customized PolyLoss for the end-to-end neural network models using multi-continuous image frames. The masked sequential autoencoders are adopted to pre-train the neural network models with reconstructing the missing pixels from a random masked image as the objective. Then, in the fine-tuning segmentation phase where lane detection segmentation is performed, the continuous image frames are served as the inputs, and the pre-trained model weights are transferred and further updated using the backpropagation mechanism with customized PolyLoss calculating the weighted errors between the output lane detection results and the labeled ground truth. Extensive experiment results demonstrate that, with the proposed pipeline, the lane detection model performance on both normal and challenging scenes can be advanced beyond the state-of-the-art, delivering the best testing accuracy (98.38%), precision (0.937), and F1-measure (0.924) on the normal scene testing set, together with the best overall accuracy (98.36%) and precision (0.844) in the challenging scene test set, while the training time can be substantially shortened.

*Index Terms*— Lane Detection, Self Pre-training, Masked Sequential Autoencoders, PolyLoss, Deep Neural Network

## I. INTRODUCTION

Lane detection is one of the crucial parts of automated driving and is the foundation of many intelligent and advanced driving assistant systems. However, lane detection has always been a challenging task, for complex and variable realistic road conditions (these scenes are easily disturbed by factors including shadows, degraded road signs, blocking, poor lighting, and bad weather), and the curved and elongated features of lane lines [1].

In recent years, many deep learning models have been proposed for vision-based lane detection [2]. Before the emergence of deep learning, traditional methods mainly utilize traditional computer vision techniques, which rely on manually manipulated operators to extract handcrafted features, including geometry [3], [4], color [5], etc., to do the detection, and then refine the results using a series of fitting methods, such as Hough transform [6] and B-spline fitting [7]. Although some progress had been made, traditional methods are not robust to complex and challenging traffic scenes. In contrast, deep learning based methods can extract more favorable features automatically and achieve superior performance in a variety of complex environments [2]. Generally, deep learning approaches are currently developed from three main perspectives: segmentation-based [1], [8]–[17], anchor-based [18]–[21], and parameter-based [22], [23], among which the most commonly used approach is the segmentation-based method. The performance of segmentation-based methods for lane detection has been continuously improving with various neural network structures developed. Getting rid of dense layers, fully Convolutional Networks (FCNs) [12], [24] employ solely locally connected layers, e.g., convolution, pooling, and upsampling, to enable efficient learning of inputs images with arbitrary sizes, which makes it well-suited for the varying input images of lane detection. Spatial convolutional neural network (SCNN) [8] adopts customized spatial convolutional layers using slice-by-slice convolutions for message passing to capture essential spatial information and correlation for lane detection. UNet-based [1], [9]–[11], [17], [25] neural networks with symmetrical encoder-decoder structures, can extract features at multiple scales, leading to accurately identifying lane markings of different sizes and shapes. Using similar symmetrical encoder-decoder structures, SegNet-based [26]–[28] models employ pooling indices for upsampling, reducing trainable parameters and memory requirements. Generative Adversarial Neural Network (GAN) [29] with embedding loss can preserve label-resembling qualities and improve the outputs' realism and structure preservation, reducing the need for complex post-processing in lane detection.

Manuscript submitted for review on August 11th, 2022, revised on May 21st, 2023, accepted on July 26th, 2023. This work was supported by the Applied and Technical Sciences (TTW), a subdomain of the Dutch Institute for Scientific Research (NWO) through the Project Safe and Efficient Operation of Automated and Human-Driven Vehicles in Mixed Traffic (SAMEN) under Contract 17187. (*Corresponding author: Yongqi Dong*).

#These authors contributed equally to this work and should be considered as co-first authors.
 R. Li is with the Department of Civil and Environmental Engineering, Villanova University, Villanova, PA 19085, USA (rli04@villanova.edu).
 Y. Dong is with the Faculty of Civil Engineering and Geosciences, Delft University of Technology, Delft, 2628 CN, The Netherlands (y.dong-4@tudelft.nl).

On the other hand, self-supervised learning has shown in recent studies [30]–[33] that learning a generic feature representation by self-supervision can enable the downstream tasks to achieve highly desirable performance. The basic idea, masking and then reconstructing, is to input a masked set of image patches to the neural network model and then reconstruct the masked patches at the output, allowing the model to learn more valuable features and aggregate contextual information. When it comes to vision-based lane detection, self-supervised learning can provide stronger feature characterization by exploring interrelationships between lane lines and other regions of the images in the continuous frames for the downstream lane detection task. With self-supervised pre-training, it is also possible to accelerate the model convergence in the training phase reducing training time. Meanwhile, with the aggregated contextual information and valuable features by pre-training, the lane detection results can be further advanced.

In this paper, a self pre-training paradigm is investigated for boosting the lane detection performance of the end-to-end encoder-decoder neural network using multi-continuous image frames. The masked sequential autoencoders are adopted to pre-train the neural network model by reconstructing the missing pixels from a randomly masked image with mean squared error (MSE) as the loss function. The pre-trained model weights are then transferred to the fine-tuning segmentation phase of the per-pixel image segmentation task in which the transferred model weights are further updated using backpropagation with a customized PolyLoss calculating the weighted errors between the output lane detection results and the labeled ground truth. With this proposed pipeline, the model performance for lane detection on both normal and challenging scenes is advanced beyond the state-of-the-art results by considerable ratios.

The main contributions of this paper are as follows:

1. This study proposes a robust lane detection pipeline through self pre-training with masked sequential autoencoders and fine-tuning with customized PolyLoss, and verified its effectiveness by extensive comparison experiments;

2. A customized PolyLoss is developed and adopted to further improve the capability of the neural network model. Without many extra parameter tuning, the customized PolyLoss can bring a significant improvement in the lane detection segmentation task while substantially accelerating model convergence speed and reducing the training time;

3. The whole pipeline is tested and verified using three deep neural network structures, i.e., UNet_ConvLSTM [1], SCNN_UNet_ConvLSTM [10], and SCNN_UNet_Attention [17], with the SCNN_UNet_Attention based model delivering the best detection results for normal testing scenes, while SCNN_UNet_ConvLSTM model delivering the best detection results for challenging scenes surpassing baseline models.

## II. PROPOSED METHOD

This study proposes a pipeline for lane detection through self pre-training with masked sequential autoencoders and fine-tuning segmentation with customized PolyLoss. In the first stage, the images are randomly masked as the inputs, and the neural network model is pre-trained with reconstructing the complete images as the objective. In the second stage, the pre-trained neural network model weights are transferred to the segmentation neural network model with the same backbone and only the structure of the output layer is adjusted. In this phase, continuous image frames without any masking are served as inputs. The neural network weights are further updated and fine-tuned by minimizing PolyLoss with the backpropagation mechanism. In this study, three neural network models, i.e., UNet_ConvLSTM [1], SCNN_UNet_ConvLSTM [10], and SCNN_UNet_Attention [17] are tested. In the last stage, post-processing methods, e.g., Density-based spatial clustering of applications with noise (DBSCAN) [34] for clustering the lane types and curve fitting to smooth the detected lines, are proposed to further improve the overall performance of the detection. However, due to time constraints and computational restrictions and following the convention in literature, e.g., [1], [10], [11], post-processing is not specifically explored in this paper. The framework of the proposed pipeline is illustrated in Fig. 1. In the remaining parts of this section, each phase will be introduced in detail.

### A. Preliminary and Network Backbone

This study tests the proposed pipeline with three hybrid neural network models based on the UNet [25] backbone, i.e, UNet_ConvLSTM [1], SCNN_UNet_ConvLSTM [10], and SCNN_UNet_Attention [17]. The three models are in similar structures composing three parts, i.e., encoder Convolutional Neural Network (CNN), Convolutional Long Short-Term Memory (ConvLSTM) block or Attention block, and decoder CNN, and they both work in an end-to-end approach.

Encoder-decoder is a widely used framework in the field of deep learning with various network structures. It is capable of mapping directly from the original input to the desired output in an end-to-end manner and keeping the input and output of the same size. Such a framework has demonstrated good performances in natural language processing tasks, e.g., machine translation, summary extraction, and computer vision tasks, e.g., target detection, scene perception, and image segmentation e.g., [1], [10], [11], [25]. Lane detection as a typical image semantic segmentation or instance segmentation task can surely be tackled with super results under the encoder-encoder structure, e.g., [1], [10], [11], [17].

A commonly used base neural network backbone for lane detection (and also other image segmentation task) is the UNet [25], which is an improved FCN. UNet with a symmetric encoder-decoder structure is originally developed to solve the problem of medical image segmentation. In UNet, a block of its encoder contains two convolutional layers, and the feature map is downsampled using pooling layers to reduce the feature map size and increase the number of channels. The decoder, which is symmetric with the encoder, performs deconvolution and upsampling operation for feature recovery and data reconstruction. The decoder CNNs have the same size and number of feature maps as in the encoder but are arranged in the opposite direction, and the feature maps are appended in a direct manner. With symmetrical CNN-based encoder-decoder structure, UNet is widely used in various aspects of segmentation tasks, including lane detection, with outstanding performances.

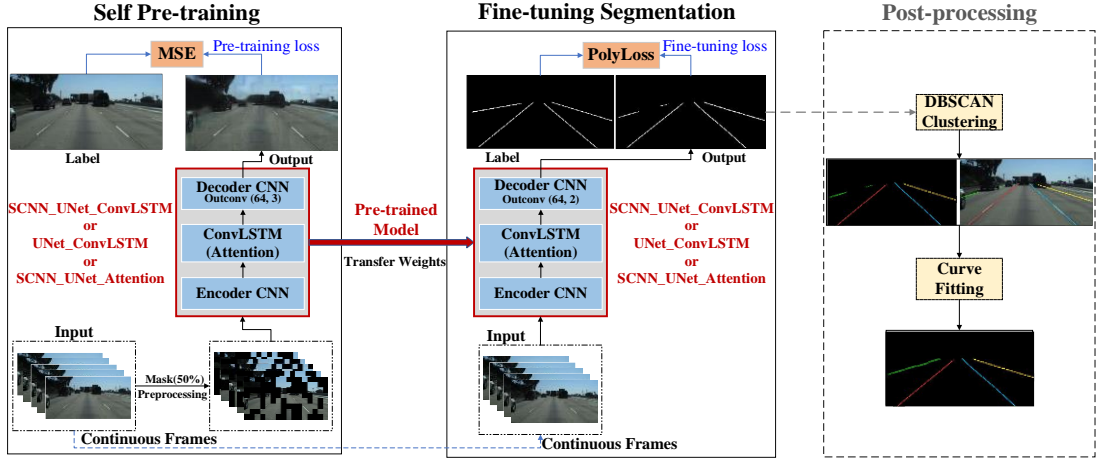

**Fig. 1.** The framework of the proposed pipeline

However, the original pure UNet does not consider the slender spatial structure and the correlations and continuity of lane lines in continuous image frames. To tap the temporal continuity of the lane line detection, the ConvLSTM module is embedded between the encoder-decoder in the UNet_ConvLSTM model [1], which can integrate the time series features extracted from the input multi-continuous frames. To further improve lane detection results, SCNN_UNet_ConvLSTM [10] incorporates SCNN in its single image feature extraction module to make use of the spatial correlations of lane structure and achieves state-of-the-art performance. SCNN_UNet_Attention [17] which applies a spatial-temporal attention module with liner LSTM in the middle of the encoder and decoder rather than ConvLSTM, can further exploit spatial-temporal correlations and dependencies of different image regions among different frames in the continuous image sequence, and further advance the detection performance. This study implemented and tested UNet_ConvLSTM, SCNN_UNet_ConvLSTM, and SCNN_UNet_Attention models to verify the proposed pipeline.

*B. Self Pre-training with Masked Sequential Autoencoders*

For vision-based lane detection, in most of the driving scene image frames, lane lines only account for a small fraction of the whole image, which means there is more spatial redundancy compared to other segmentation tasks. It is vital but challenging to make full use of the valuable features and aggregate contextual information, especially the interrelationships between lane lines and other regions of the images in continuous frames.

He et al. [31] show that taking advantage of a pre-training strategy by randomly masking a high proportion of input image and reconstructing the original image from the masked patches using the latent representations can improve accuracy and accelerate training speed for downstream tasks. That is, the images with a high masking rate are input into the designed model for reconstruction as a self-supervised learning task, and then the pre-trained model can be migrated to the downstream tasks for fine-tuning. With this pre-training method, the model can gain a better overall "understanding" of the images, since reconstructing the masked pixels in the pre-training phase facilitates the trained model a good generalization capability, which can serve for downstream tasks.

Inspired by and upgraded upon the idea of self-training by "random masking-reconstructing" with autoencoders [31], this study proposed to incorporate a pre-training phase with masked sequential autoencoders to pre-train the lane detection models and facilitate their capabilities in aggregating contextual information for feature extraction through continuous frames. In the pre-training phase, $S$ (for the experiments carried out in this study, $S = 5$) consecutive images are used as the inputs with every image getting certain parts randomly masked. To implement the masking, each of the input images with the size of (128×256) is firstly divided into non-overlapping patches with the size of (16×16), and then random masking is applied to mask a certain ratio of the (8×16=128) patches in each image. The original last image within the input consecutive five image frames is set as the target of the reconstruction task. Using the mean squared error (MSE) as the loss function, the image reconstruction task can be expressed as a minimization problem by (1):

$$\min \quad \frac{1}{S}\sum_{k=1}^{S} d_2(M_k, P_k) \qquad (1)$$

where $S$ is the number of image samples; $M_k$ is the pixel value matrix with a size of (128×256) containing all pixel values in the reconstructed image $k$ reconstructed from the one with masked patches; $P_k$ is the pixel value matrix with a size of (128×256) containing all pixel values in the original image $k$; $d_2(\cdot)$ means Euclidean norm which calculates the Euclidean distance between the matrix $M_k$ and $P_k$, and can be illustrated by (2):

$$d_2(M_k, P_k) = \frac{1}{h*w}\sum_{i=1}^{h}\sum_{j=1}^{w}(m_{i,j} - p_{i,j})^2 \qquad (2)$$

where $m_{i,j}$ and $p_{i,j}$ are the pixel values on $i^{th}$ row $j^{th}$ column in the constructed image matrix $M_k$ and original image matrix $P_k$ respectively; $h$ is the height of the image with $h=128$ in this study; $w$ is the width of the image with $w=256$ in this study.

Using UNet_ConvLSTM, SCNN_UNet_ConvLSTM, and SCNN_UNet_Attention models, the input continuous images with maskings are downsampled four times consecutively by the encoder, and the extracted time-series features of size (8×16×512) are then transferred to the ConvLSTM module (or Attention module) for spatial-temporal features integration. Finally, the decoder upsamples the integrated features four times into the same size as the input image and calculates the MSE loss between the reconstructed 5$^{th}$ image and the original 5$^{th}$ image of the input frames. Note in the pre-training phase, the output layers of both UNet_ConvLSTM, SCNN_UNet_ConvLSTM, and SCNN_UNet_Attention, are adjusted from the original models reported in [1], [10], and [17], with the number of channels changed to 3 (check Fig. 2).

Regarding the masking ratio, the results of ablation tests with ratios set at 25%, 50%, or 75% found that a 50% ratio delivers a balanced performance. Thus, in the pre-training phase, the random masking ratio is set at 50% for all models.

Different from the original masked autoencoders [31] implemented by the vision transformer, the proposed upgrade version of masked sequential autoencoders for pre-training is implemented under the "CNN-ConvLSTM-CNN" or "CNN-Attention_LSTM-CNN" architecture, which can further aggregate valuable image contextual information and spatial-temporal features. By masking the whole continuous 5 image frames and only recovering the last frame, which is also the current frame for lane detection, the proposed upgraded masked sequential autoencoders facilitate the model to learn not only correlations of different regions within one image but also the spatial-temporal interrelationships and dependencies between different regions of the images among continuous frames.

*C. Fine-tuning with PolyLoss*

Vison-based lane detection as a typical segmentation task aims to classify the image at the pixel level, labeling each pixel with its corresponding class, i.e., lane or background. Generally, for a segmentation task, the input is one image, but in the proposed pipeline, a continuous image sequence is used as input, and only the last image of the continuous sequence is segmented, check Fig 1 for details.

By pre-training with reconstructing the masked patches, the pre-trained model should already get the aggregate contextual information and valuable spatial-temporal features, however, fine-tuning is required to further train the model to adapt it to the per-pixel segmentation task, making full use of those extracted features.

With the elongated structure, lane lines often occupy only a very small fraction of the overall pixels in an image, making lane detection a typical imbalanced two-class segmentation task. Usually, weighted cross entropy loss is adopted for addressing this imbalanced two-class segmentation, which reshapes the standard cross entropy (CE) loss by introducing weighting factors to reduce the weights of the background samples and focus more on the weights of lane pixels. However, literature [35], [36] revealed that weighted CE loss does not perform well under certain situations with severely imbalanced data. To further improve the performance of the lane detection models and improve the capabilities of handling the severe imbalance between lane line and background pixels, this study customizes a PolyLoss (PL for short in the model names), and tests and verifies its effectiveness.

PloyLoss is based on the Taylor expansions of CE loss and focal loss (FL), which treats the loss functions as a linear combination of polynomial functions [36]. The CE loss and FL loss can be expressed in (3) and (4):

$$L_{CE} = -\log(Q_t) \tag{3}$$

$$L_{FL} = -\alpha(1-Q_t)^\varepsilon \log(Q_t) \tag{4}$$

where $L_{CE}$ and $L_{FL}$ stand for the CE loss and FL loss respectively; $Q_t$ is the prediction probability of the target ground-truth class; $\alpha, \varepsilon$ are the tunable hyperparameters for $L_{FL}$.

The loss functions of both CE and FL can be decomposed into a series of weighted polynomial bases in the form of $\sum_{j=1}^{\infty} \alpha_j (1-Q_t)^j$ where $j \in \mathbb{Z}^+$, $\alpha_j \in R^+$ is the polynomial coefficient. Each polynomial basis $(1-Q_t)^j$ is weighted by the corresponding polynomial coefficients $\alpha_j \in R^+$, so that it is easy to adjust the different polynomial bases of PolyLoss. The Taylor expansion of FL, indicated by $L_{FL-T}$, is given in (5):

$$L_{FL-T} = -(1-Q_t)^\varepsilon \log(Q_t) = \sum_{j=1}^{\infty} \frac{(1-Q_t)^{j+\varepsilon}}{j}$$

$$= (1-Q_t)^{1+\varepsilon} + \frac{(1-Q_t)^{2+\varepsilon}}{2} + ... + \frac{(1-Q_t)^{N+\varepsilon}}{N} + \frac{(1-Q_t)^{N+1+\varepsilon}}{N+1} + ... \tag{5}$$

where $N \in \mathbb{Z}^+$; $\varepsilon$ is a modulating factor, with which the FL can simply shift the power $j$ by $\varepsilon$, i.e., shift all polynomial coefficients horizontally by $\varepsilon$ [36].

To improve the model performance and robustness, dropping the higher order polynomials and tuning the leading polynomials are applied in previous studies [36], [37]. Similarly here, after truncating all higher order $(N+1 \rightarrow \infty)$ polynomial terms and tuning the leading $N$ polynomials using the perturbation term $\gamma_j$, $j=1,2,3,\cdots,N$, the truncated $L_{PL-N}$ is obtained and shown in (6):

$$L_{PL-N} = (\gamma_1+1)(1-Q_t)^{1+\varepsilon} + (\gamma_2+\frac{1}{2})(1-Q_t)^{2+\varepsilon} + ... + (\gamma_N+\frac{1}{N})(1-Q_t)^{N+\varepsilon}$$

$$= -(1-Q_t)^\varepsilon \log(Q_t) + \sum_{j=1}^{N} \gamma_j (1-Q_t)^{j+\varepsilon} \tag{6}$$

To further simplify the $L_{PL-N}$ and render it applicable to be easily tuned for different tasks and data sets, Leng et.al [36] carried out extensive experiments and observed that adjusting a single coefficient for the leading polynomial can achieve better performance than the original FL loss. With this, the general form of the finally simplified formula of PolyLoss $L_{PL}$ (of FL) is illustrated by (7):

$$L_{PL} = -\alpha(1-Q_t)^\varepsilon \log(Q_t) + \gamma(1-Q_t)^{\varepsilon+1} \quad (7)$$

where $\alpha, \gamma, \varepsilon$ are the tunable hyperparameters. Adapting it to the imbalanced two-class segmentation task of lane detection, this study further customized (7) into (9) which will be discussed in the following *subsection E*.

More details about PolyLoss can be referred to in [36].

*D. Postprocessing Phase*

Since in real driving scenarios, it is necessary to identify the types and colors of the lane lines (e.g., dashed lines, continuous double yellow lines), the detected lane lines need to be grouped into different colors to indicate their different types, i.e., lane detection considered as an instance segmentation task. With the fine-tuning lane line segmentation outputs, the DBSCAN [34] algorithm is proposed to cluster the detected lane lines to diffident colors, indicating different types. Then, curve fitting is proposed at the end to smooth the detected lines repairing the discontinuous broken ones (see the post-processing section in Fig. 1). One needs to note that this paper only presents here the idea of post-processing which can serve to upgrade the lane detection results, however, all the results in this paper do not use post-processing which follows the general convention in literature, e.g., [1], [10], [11], [17].

*E. Implementation Details*

**Configuration Details**: In this paper, to reduce the computational payload and save training time, the size of the images for both the training set and test set is set to a resolution of 128×256. In pre-training, the proportion of masked patches is set to 50%. Experiments were carried out on two NVIDIA Tesla V100 (32 GB memory) GPUs, using PyTorch version 1.9.0 with CUDA Deep Neural Network library (cuDNN) version 11.1. The batch size is set to be as large as possible, which is 60. The learning rate was initially set to 0.001 with decay applied after each epoch.

**Network Details**: In network models of UNet_ConvLSTM, SCNN_UNet_ConvLSTM, and SCNN_UNet_Attention, most of the convolutional kernel size is 3×3, except for the SCNN block in SCNN_UNet_ConvLSTM and SCNN_UNet_Attention. The encoder part (see the left *Encoder* section in Fig. 2) uses two convolutional layers as a downsampling block, in which the size of the feature map is reduced by half and the number of channels is doubled by the pooling layer. Four successive downsampling blocks are performed, and the last downsampling block does not change the number of output channels compared with its input. The final feature map of the encoder with a size of 8×16×512 is fed into the spatial-temporal ConvLSTM (or Attention) module.

The sequential features of the feature map are learned in the ConvLSTM/Attention module, which is equipped with 2 hidden layers of size 512 and outputs an 8×16 feature map of the same size as its input. The decoder network (check the *Decoder* part in Fig. 2), is with the same size and number of feature maps as in the encoder but of a reverse-arranged symmetric structure that upsamples the extracted features to the original size of the input image. One needs to note that, in the pre-training task, to recover the image into original RGB pixels, the number of channels in the output layer of the decoder is set as 3; while in the fine-tuning segmentation phase, it is set as 2 for the two-class segmentation task. Therefore, for model weights transfer from the pre-training to fine-tuning phase, the pre-trained model weights are transferred to the fine-tuning model except for the weights of the output layer. Both the pre-training and fine-tuning segmentation phases output images of the same size as the input one. Details can be checked in Fig. 2.

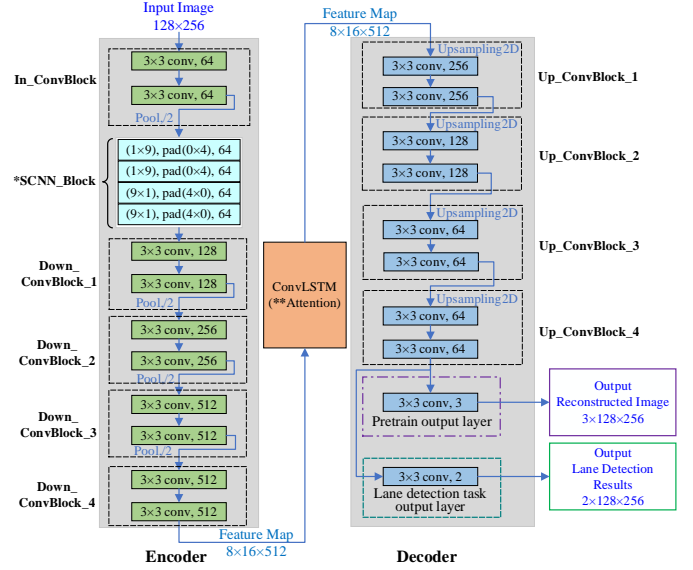

**Fig. 2.** Pre-training network and lane line detection neural network structure.

*\*SCNN block is for SCNN_UNet_ConvLSTM and SCNN_UNet_Attention, UNet_ConvLSTM does not have it.*
*\*\*Attention block is only for SCNN_UNet_Attention model.*

**Loss Function Details:** As mentioned before, to make the proposed pipeline work, different loss functions are adopted accordingly in different phases. In the pre-training phase, the objective is to reconstruct the masked images, and for that, the mean square error (MSE) is selected as the loss function.

While in the fine-tuning segmentation phase, the objective is to segment the pixels into lanes or backgrounds, which is a typical discriminative binary segmentation task. This study tested the weighted cross entropy loss and the customized PolyLoss and compared their performances in tackling the imbalanced lane segmentation task. The two tailored losses applied in the fine-tuning segmentation phase are illustrated by (8) and (9).

$$CE = -\frac{1}{T}\sum_{i=1}^{T}[\omega_1 y_i \log(h_\theta(x_i)) + \omega_0(1-y_i)\log(1-h_\theta(x_i))] \quad (8)$$

$$PL = -\frac{1}{T}\sum_{i=1}^{T}\left(\alpha\begin{bmatrix}y_i(1-h_\theta(x_i))^\varepsilon \log(h_\theta(x_i)) + \\ (1-y_i)h_\theta(x_i)^\varepsilon \log(1-h_\theta(x_i))\end{bmatrix} - \gamma\begin{bmatrix}y_i(1-h_\theta(x_i))^{\varepsilon+1}) + \\ (1-y_i)h_\theta(x_i)^{\varepsilon+1}\end{bmatrix}\right) \quad (9)$$

where CE and PL stands for the weighted cross entropy loss and the customized PolyLoss, respectively; $T$ is the number of training examples; $y_i$ is the true segmentation label for the training example $i$; $\omega_1$ and $\omega_0$ stands for the weights for lane class and background class respectively; $x_i$ is the input training example $i$; $h_\theta(\cdot)$ represents the neural network model with trainable weights $\theta$; and $\alpha$, $\gamma$, $\varepsilon$ are the tunable hyperparameters for the customized PolyLoss, which are determined by grid search method.

**Optimizer Details:** To efficiently train and validate the proposed model pipeline, different optimizers were tested in different stages. Three optimizers, Stochastic Gradient Descent (SGD), Adaptive Moment Estimation (Adam), and Rectified Adaptive Moment Estimation (RAdam), were tested in the pre-training and fine-tuning segmentation phases. Compared to Adam, SGD requires more parameters, decreases more slowly, and may oscillate continuously on both sides of the gully. Through the tests, Adam performed better than SGD in both the pre-training task and the fine-tuning lane segmentation task. Furthermore, RAdam solves the problem of falling into local optimization that is easily encountered by Adam, and is more robust to the changes of learning rate. Experiments verified that there was even a slight improvement in the model performance of RAdam over Adam. Therefore, the RAdam optimizer was finally chosen for both the pre-training and the fine-tuning segmentation phases.

## III. EXPERIMENTS AND RESULTS

### A. Datasets Descriptions

To verify the proposed pipeline, a lane image dataset with continuous image frames is required. Although there are various open-sourced lane detection image datasets, e.g., CULane [8], CurveLane [38], seldom do they contain continuous frames. Therefore, this study adopted the tvtLANE [1] dataset, which is upgraded on the TuSimple lane detection challenge dataset, to train and verify the proposed method. There are one integrated training dataset and two testing sets in tvtLANE.

The tvtLANE dataset is mainly built based on the TuSimple lane detection challenge dataset. In the original TuSimple dataset, there are 3,626 training segments and 2,782 test segments with 20 continuous frames in each segment. The images are collected in different scenes at different times, and only the last frame of each segment, e.g., the 20th frame, is labeled with ground truth. Zou et al. [1] additionally labeled the 13th image in each segment and enlarged the dataset by adding 1,148 segments of rural driving scenes collected in China. Furthermore, data augmentation methods with cropping, flipping, and rotating operations are employed, and finally a total number 19,096 continuous segments are produced.

The tvtLANE consists of two test sets, i.e., test set #1 (normal) which is built on the original TuSimple test set for normal driving scenario testing, and test set #2 (challenging) which consists of 12 challenging driving scenarios for robustness evaluation. More details of tvtLANE can be found in [1], [10].

In this study, 5 images are sampled from the continuous frames with a fixed stride. The sampling strides and frames used in the training and testing sets are elaborated in Table I, and image samples are demonstrated in Fig. 3.

TABLE I
SAMPLE METHODS FOR THE TRAINSET AND TESTSET

| Subset | Labeled Ground Truth | Sample Stride | Sample Frames |
|---|---|---|---|
| Trainset | 13th | 3 | 1st,4th,7th,10th,13th |
| | | 2 | 5th,7th,9th,11th,13th |
| | | 1 | 9th,10th,11th,12th,13th |
| | 20th | 3 | 8th,11th,14th,17th,20th |
| | | 2 | 12th,14th,16th,18th,20th |
| | | 1 | 16th,17th,18th,19th,20th |
| Test set #1 *Normal* | 13th | 1 | 9th,10th,11th,12th,13th |
| | 20th | 1 | 16th,17th,18th,19th,20th |
| Test set #2 *Challenging* | All | 1 | 1st,2nd,3rd,4th,5th |
| | | | 2nd,3rd,4th,5th,6th |
| | | | 3rd,4th,5th,6th,7th |
| | | | … |

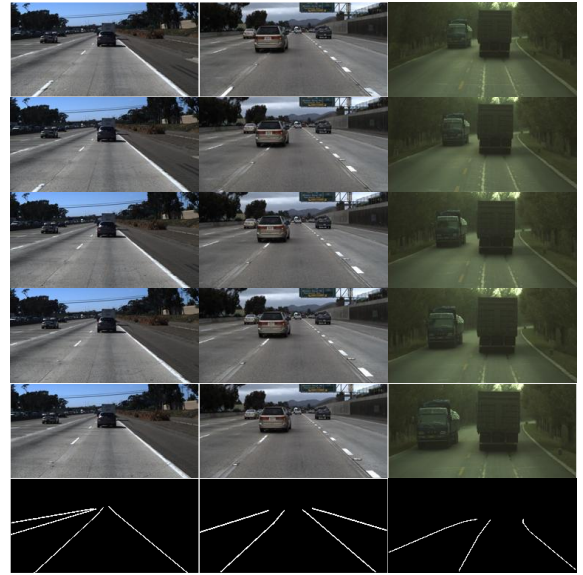

**Fig. 3.** Image samples in the tvtLANE training set and the test set. The first five images in each column are the inputs of consecutive frames, and the sixth one is the labeled ground truth of the last image in the consecutive frames. The first column is one sample in the training set, the second column is for the test set #1 (normal), and the third column is for test set #2 (challenging).

## B. Evaluation Metrics

Overall, the model performance is evaluated in terms of both visual qualitative examination with results demonstration and quantitative analysis with metrics. Considering the vision-based lane detection task as a pixel-level classification task, commonly used metrics, i.e., accuracy, precision, recall, and F1-measure [1], [10], [11], [17], are adopted. The calculations of these metrics are illustrated by (10)-(13).

$$Accuracy = \frac{Truly\ Classified\ Pixels}{Total\ Number\ of\ Pixels} \quad (10)$$

$$Precision = \frac{True\ Positive}{True\ Positive + False\ Positive} \quad (11)$$

$$Recall = \frac{True\ Positive}{True\ Positive + False\ Negative} \quad (12)$$

$$F1-measure = 2*\frac{Precision*Recall}{Precision+Recall} \quad (13)$$

In the above equations, true positive indicates the number of image pixels that are lane lines and are correctly identified; False positive indicates the number of image pixels that are background but incorrectly classified as lane lines; False negative indicates the number of image pixels that are lane lines but incorrectly classified as background.

Furthermore, for estimating the models' computational complexities, the model parameter size, i.e., Params (M), and the multiply-accumulate (MAC) operations, i.e., MACs (G), are provided.

## C. Results

In this sub-section, reconstruction performance in the self pr-training phase will be visually demonstrated, while the lane detection testing results of various models on both tvtLANE test set#1 (normal) and tvtLANE test set#2 (challenging) will be evaluated qualitatively and quantitatively.

**Self pre-training results:** Fig. 4 shows the reconstructing results of the masked images in the pr-training phase. It can be seen that the masked patches in the images can be restored very well. Although there are some minor blurs in certain images, the reconstructed images generally recover the main and critical patterns.

**Testing results on tvtLANE testset#1 (normal)**: Fig. 5, Fig. 7 (A), and Table. II (a) demonstrate the qualitative and quantitative testing results on tvtLANE testset#1 (normal).

Qualitatively, for the lane detection segmentation task, the model should be able to accurately predict the total number of lane lines, correctly detecting the location of the lane lines while avoiding unexpected broken lines and blurs. Visualizations of the lane detection results show that models using the proposed self pre-training method generally perform better than those without. Furthermore, models using the customized PolyLoss generally outperform those using weighted cross entropy loss with thinner detected lane lines and fewer blurs. Aligning with previous studies [1], [10], [11], [17], models using multi-continuous image frames defeat those using one single image as indicated in rows (c) and (d) there are fatter lane lines, merged lanes, and blurred areas at the top boundary of the image, and even wrongly detected lane numbers (check the first column in Fig. 7 (A)). One can also notice that even when vehicles or shadings of the vehicles are blocking the lane lines, the models with the proposed pretraining method and using the proposed PolyLoss can identify the lane lines completely and continuously with correct locations (check the first, fourth, and sixth columns in Fig. 7 (A)), which is crucial for vehicle localization.

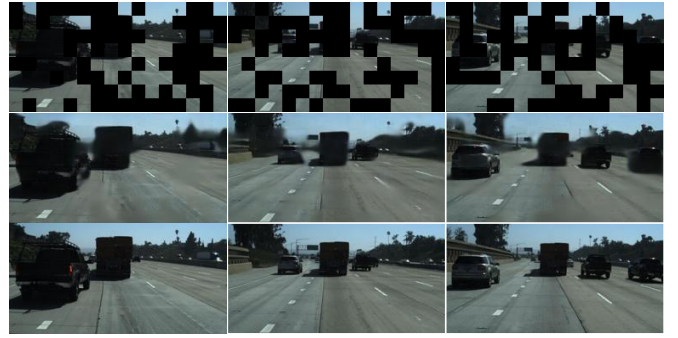

**Fig. 4.** Visualization of the reconstructing results in the pre-training phase. The first row shows images with 50% of the patches masked. The second row shows the reconstructed images after pre-training. The third row shows the original images.

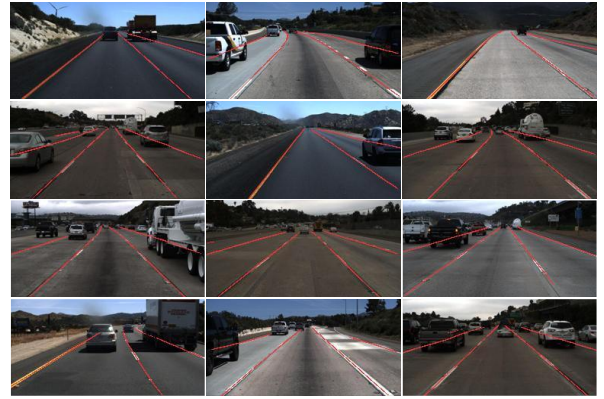

**Fig. 5**. Lane detection results obtained by SCNN_UNet_Attention_PL$^{**}$ on tvtLANE test set #1 (normal) without post-processing.

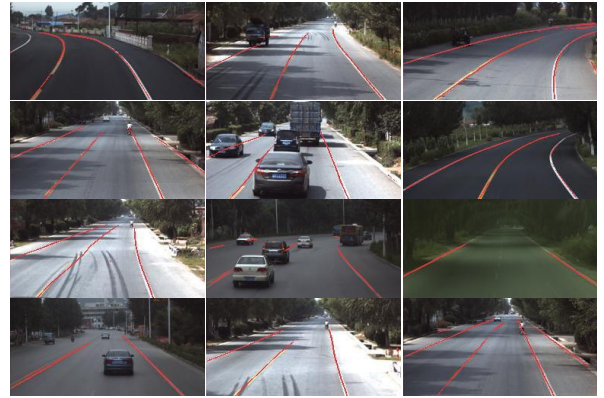

**Fig. 6** Lane detection results obtained by SCNN_UNet_ConvLSTM_PL$^{**}$ on tvtLANE test set #2 (challenging) without post-processing.

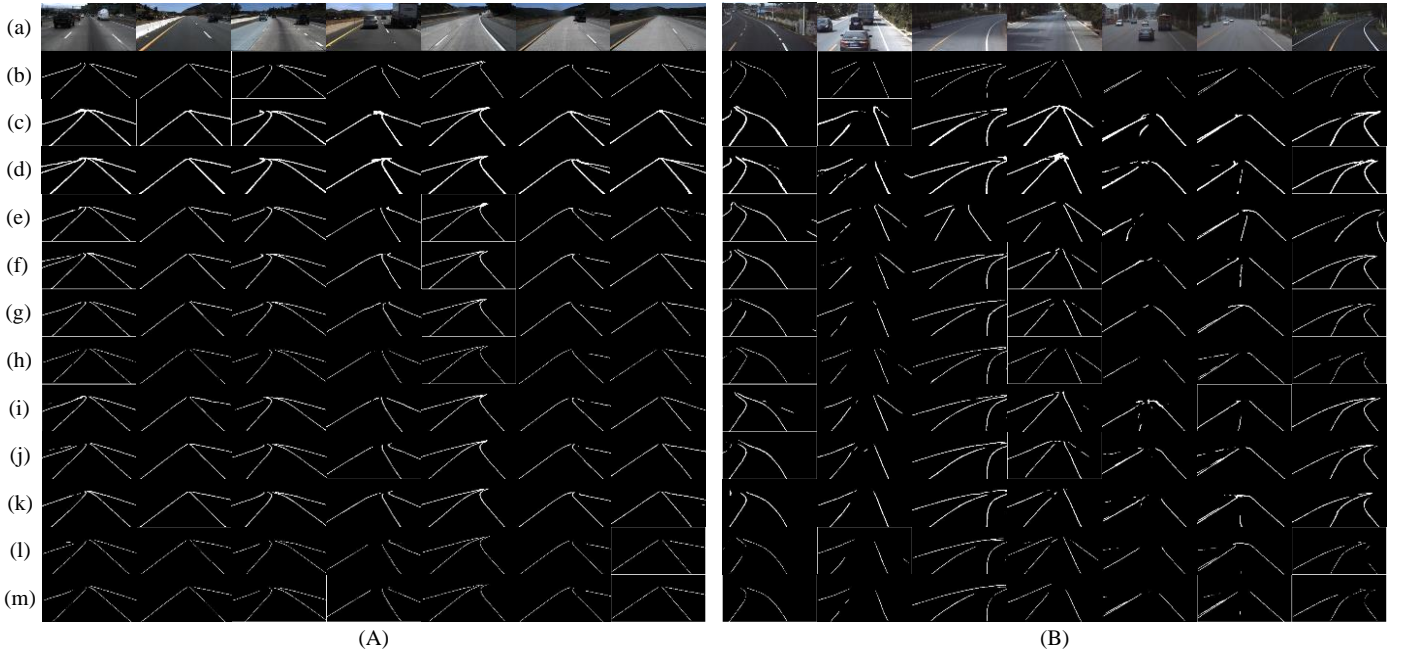

**Fig. 7** Qualitative visual comparison of the lane detection results testing on tvtLANE test set #1 (normal) (A) and tvtLANE test set #2 (challenging) (B). All results in the figure are without post-processing. (a) Original input images; (b) Ground truth; (c)~(l) are the lane detection results corresponding to the models: (c) SegNet, (d) UNet, (e) SegNet_ConvLSTM [1], (f) UNet_ConvLSTM, (g) UNet_ConvLSTM_CE**, (h) UNet_ConvLSTM_PL**, (i) SCNN_SegNet_ConvLSTM [10], (j) SCNN_UNet_ConvLSTM, (k) SCNN_UNet_ConvLSTM_ CE**, (l) SCNN_UNet_ConvLSTM_PL**, (m) SCNN_UNet_Attention_PL**. (Note: CE and PL are short for weighted cross entropy loss and PolyLoss respectively, while ** means the model is pre-trained with the proposed self pre-training method.)

Quantitatively, Table. II (a) demonstrates that the proposed self pre-training method improves the lane detection results for both UNet_ConvLSTM and SCNN_UNet_ConvLSTM models and the models using the customized PolyLoss all outperform those using the weighted cross entropy loss regarding the accuracy, precision, and F1-measure. To be specific, with the self pre-training pipeline and using the customized PolyLoss, UNet_ConvLSTM_PL** advances a lot from the baseline UNet_ConvLSTM with testing accuracy improved from 98.00% to 98.34%, precision improved from 0.857 to 0.921, and F1-measure improved from 0.904 to 0.915; while SCNN_UNet_ConvLSTM_PL** also improves a lot from the baseline SCNN_UNet_ConvLSTM with testing accuracy improved from 98.19% to 98.38%, precision improved from 0.889 to 0.929, and F1-measure improved from 0.918 to 0.922. All the models' parameter sizes and MACs do not increase.

One can find that for both models the most significant improvement was identified in precision, (i.e., 0.857 to 0.921 and 0.889 to 0.929). The higher the precision the lower the false positive is (check (11)) which means the models become more strict on pixel samples to be classified as the lane line contributing to fewer wrong detected lane pixels, which is also illustrated by the thinner detected lane lines in Fig. 7 (A). However, this might increase the number of lane pixels that are incorrectly identified as background, i.e., higher false negative, thus the recall ratio decreases. Therefore, the F1-measure which balances precision and recall, is a more reasonable evaluation measure to serve as the main benchmark [1], [8], [10], [11], [17]. Furthermore, SCNN_UNet_Attention, which was tested only under the best setting of using pre-training and customized PolyLoss, obtained the best precision (0.937) and F1-measure (0.924) beating all other state-of-the-art baseline models on this tvtLANE testset#1 (normal scene testing).

**Testing results on tvtLANE test set#2 (challenging)**: Fig. 6, Fig. 7 (B), and Table II (b) demonstrate the qualitative and quantitative testing results on tvtLANE testset#2 (challenging).

Qualitatively, as illustrated in Fig. 6 and Fig. 7 (B), when testing on the challenging driving scenes, all the models do not perform well. However, the results obtained by the models using the proposed self pre-training method are still better than those without pre-training. Especially models adopting the customized PolyLoss still output thinner lanes with less blur and more correct lane numbers.

Quantitatively, as shown in Table II (b), models with pre-training generally outperform those without regarding the overall accuracy and precision. Typically, using self pre-training method plus the customized PolyLoss, the developed UNet_ConvLSTM_PL** model obtains the best overall accuracy (98.38%), and together with other proposed models

TABLE II
MODEL PERFORMANCE COMPARISON
(a) tvtLANE TEST SET #1 (NORMAL)

| | Models | Test_Acc (%) | Precision | Recall | F1-Measure | MACs(G) | Params(M) |
|---|---|---|---|---|---|---|---|
| Using single image | SegNet | 96.93 | 0.796 | 0.962 | 0.871 | 50.2 | 29.4 |
| | UNet | 96.54 | 0.790 | **0.985** | 0.877 | 15.5 | 13.4 |
| | SCNN* | 96.79 | 0.654 | 0.808 | 0.722 | 77.7 | 19.2 |
| | LaneNet* | 97.94 | 0.875 | 0.927 | 0.901 | 44.5 | 19.7 |
| Using multi-continuous images | SegNet_ConvLSTM | 97.92 | 0.874 | 0.931 | 0.901 | 217.0 | 67.2 |
| | UNet_ConvLSTM | 98.00 | 0.857 | 0.958 | 0.904 | 69.0 | 51.1 |
| | UNet_ConvLSTM_CE** | 98.19 | 0.882 | 0.940 | 0.910 | 69.0 | 51.1 |
| | UNet_ConvLSTM_PL** | 98.34 | 0.921 | 0.909 | 0.915 | 69.0 | 51.1 |
| | SCNN_SegNet_ConvLSTM | 98.07 | 0.893 | 0.928 | 0.910 | 223.0 | 67.3 |
| | SCNN_UNet_ConvLSTM | 98.19 | 0.889 | 0.950 | 0.918 | 93.0 | 51.3 |
| | SCNN_UNet_ConvLSTM_CE** | 98.20 | 0.891 | 0.952 | 0.921 | 93.0 | 51.3 |
| | SCNN_UNet_ConvLSTM_PL** | **98.38** | 0.929 | 0.915 | 0.922 | 93.0 | 51.3 |
| | SCNN_UNet_ Attention _PL** | 98.36 | **0.937** | 0.911 | **0.924** | 68.9 | 13.7 |

(b) tvtLANE TEST SET #2 (CHALLENGING)

| Models \ Challenging Scenes | overall | 1-curve&occlude | 2-shadow-bright | 3-bright | 4-occlude | 5-curve | 6-dirty&occlude | 7-urban | 8-blur&curve | 9-blur | 10-shadow&dark | 11-tunnel | 12-dim&occlude |
|---|---|---|---|---|---|---|---|---|---|---|---|---|---|
| **PRECISION** | | | | | | | | | | | | | |
| SegNet | 0.6080 | 0.6810 | 0.7067 | 0.5987 | 0.5132 | 0.7738 | 0.2431 | 0.3195 | 0.6642 | 0.7091 | 0.7499 | 0.6225 | 0.6463 |
| UNet | 0.6754 | 0.7018 | 0.7441 | 0.6717 | 0.6517 | 0.7443 | 0.3994 | 0.4422 | 0.7612 | 0.8523 | 0.7881 | 0.7009 | 0.5968 |
| SegNet_ConvLSTM | 0.7563 | 0.8176 | 0.8020 | 0.7200 | 0.6688 | 0.8645 | 0.5724 | 0.4861 | 0.7988 | 0.8378 | 0.8832 | 0.7733 | 0.8052 |
| UNet_ConvLSTM | 0.7784 | 0.7591 | 0.8292 | 0.7971 | 0.6509 | 0.8845 | 0.4513 | 0.5148 | 0.8290 | **0.9484** | 0.9358 | 0.7926 | 0.8402 |
| UNet_ConvLSTM_CE** | 0.7932 | 0.8004 | 0.8312 | 0.8285 | 0.7661 | 0.8557 | 0.5242 | 0.5567 | 0.7545 | 0.9200 | 0.9312 | 0.8496 | 0.8026 |
| UNet_ConvLSTM_PL** | 0.8331 | 0.8429 | **0.8824** | 0.8691 | 0.8125 | **0.9578** | 0.5970 | **0.5591** | 0.8289 | 0.9247 | **0.9634** | 0.7688 | 0.9160 |
| SCNN_SegNet_ConvLSTM | 0.7673 | 0.8326 | 0.7497 | 0.7470 | 0.7369 | 0.8647 | 0.6196 | 0.4333 | 0.7371 | 0.8566 | 0.9125 | 0.8153 | 0.8466 |
| SCNN_UNet_ConvLSTM | 0.7784 | 0.8182 | 0.8362 | 0.8189 | 0.7359 | 0.8365 | 0.5872 | 0.5377 | 0.8046 | 0.8770 | 0.8722 | 0.7952 | 0.7817 |
| SCNN_UNet_ConvLSTM_CE** | 0.8001 | 0.8754 | 0.8672 | 0.8519 | 0.7763 | 0.8664 | 0.5523 | 0.5261 | 0.7396 | 0.8865 | 0.8974 | 0.8115 | 0.9101 |
| SCNN_UNet_ConvLSTM_PL** | **0.8444** | 0.9074 | 0.8757 | 0.8644 | 0.8464 | 0.9049 | **0.7177** | 0.4827 | 0.8157 | 0.9440 | 0.9606 | **0.8736** | 0.9220 |
| SCNN_UNet_ Attention _PL** | 0.8413 | **0.9189** | 0.8763 | **0.8838** | 0.8598 | 0.9238 | 0.6210 | 0.5229 | **0.8847** | 0.9039 | 0.9229 | 0.8408 | **0.9369** |
| **F1-MEASURE** | | | | | | | | | | | | | |
| SegNet | 0.6727 | 0.8042 | 0.7900 | 0.7023 | 0.6127 | 0.8639 | 0.2110 | 0.4267 | 0.7396 | 0.7286 | 0.7675 | 0.6935 | 0.5822 |
| UNet | 0.6985 | 0.8200 | 0.8408 | 0.7946 | 0.7337 | 0.7827 | 0.3698 | 0.5658 | 0.8147 | 0.7715 | 0.6619 | 0.5740 | 0.4646 |
| SegNet_ConvLSTM | 0.7609 | 0.8852 | 0.8544 | 0.7688 | 0.6878 | 0.9069 | 0.4128 | 0.5317 | 0.7873 | 0.7575 | 0.8503 | 0.7865 | 0.7947 |
| UNet_ConvLSTM | 0.7143 | 0.8465 | **0.8891** | 0.8411 | 0.7245 | 0.8662 | 0.2417 | 0.5682 | 0.8323 | **0.7852** | 0.6404 | 0.4741 | 0.5718 |
| UNet_ConvLSTM_CE** | 0.6537 | 0.8365 | 0.8697 | 0.8263 | 0.7614 | 0.8165 | 0.2440 | 0.5359 | 0.7618 | 0.7206 | 0.4832 | 0.3274 | 0.2595 |
| UNet_ConvLSTM_PL** | 0.6284 | 0.8220 | 0.8731 | 0.8300 | 0.7705 | 0.8295 | 0.1845 | 0.4426 | 0.7278 | 0.5712 | 0.4157 | 0.3545 | 0.4821 |
| SCNN_SegNet_ConvLSTM | **0.7666** | 0.8956 | 0.8237 | 0.7909 | 0.7468 | **0.9108** | 0.4398 | 0.4858 | 0.7379 | 0.7546 | **0.8729** | 0.7963 | 0.8074 |
| SCNN_UNet_ConvLSTM | 0.7024 | 0.8670 | 0.8866 | 0.8405 | 0.7565 | 0.7955 | 0.4179 | **0.5933** | 0.7880 | 0.7285 | 0.6296 | 0.4747 | 0.4134 |
| SCNN_UNet_ConvLSTM_CE** | 0.7327 | 0.8937 | 0.8690 | **0.8426** | 0.7656 | 0.8352 | 0.2493 | 0.5751 | 0.7756 | 0.7122 | 0.7661 | 0.6989 | 0.5420 |
| SCNN_UNet_ConvLSTM_PL** | 0.6711 | 0.8685 | 0.8796 | 0.8161 | **0.7988** | 0.7897 | 0.2853 | 0.4921 | 0.8258 | 0.7255 | 0.5244 | 0.3963 | 0.3255 |
| SCNN_UNet_ Attention _PL** | 0.6772 | 0.8530 | 0.8771 | 0.8111 | 0.7579 | 0.7881 | 0.2926 | 0.5057 | **0.8595** | 0.7569 | 0.5857 | 0.3737 | 0.4565 |
| **ACCURACY (%)** | | | | | | | | | | | | | |
| SegNet | 96.57 | 96.72 | 96.16 | 96.01 | 96.83 | 96.50 | 95.93 | 96.16 | 96.39 | 96.12 | 97.26 | 96.79 | 97.37 |
| UNet | 96.68 | 96.68 | 96.00 | 95.78 | 97.06 | 96.35 | 95.45 | 96.35 | 96.58 | 96.62 | 97.50 | 97.53 | 97.58 |
| SegNet_ConvLSTM | 97.83 | 98.10 | 97.38 | 97.52 | 98.17 | 97.72 | 96.98 | 97.92 | 97.61 | 97.08 | 98.39 | 98.07 | 98.26 |
| UNet_ConvLSTM | 97.93 | 97.83 | 97.48 | 97.70 | 97.94 | 97.73 | 97.27 | 97.86 | 97.75 | 97.65 | 98.49 | 98.37 | 98.38 |
| UNet_ConvLSTM_CE** | 98.13 | 98.19 | 97.72 | 98.04 | 98.47 | 97.77 | 97.41 | 98.30 | 97.67 | 97.69 | 98.58 | 98.54 | 98.57 |
| UNet_ConvLSTM_PL** | **98.38** | 98.60 | **98.06** | 98.33 | 98.75 | **98.35** | 97.66 | **98.61** | 98.09 | 97.77 | 98.63 | **98.63** | 98.63 |
| SCNN_SegNet_ConvLSTM | 97.90 | 98.24 | 97.21 | 97.68 | 98.39 | 97.73 | 97.11 | 97.80 | 97.48 | 97.29 | 98.50 | 98.28 | 98.34 |
| SCNN_UNet_ConvLSTM | 97.95 | 98.08 | 97.45 | 97.86 | 98.31 | 97.63 | 97.17 | 97.95 | 97.63 | 97.43 | 98.41 | 98.39 | 98.39 |
| SCNN_UNet_ConvLSTM_CE** | 98.03 | 98.33 | 97.64 | 98.05 | 98.45 | 97.69 | 97.42 | 97.95 | 97.54 | 97.57 | 98.38 | 98.23 | 98.56 |
| SCNN_UNet_ConvLSTM_PL** | 98.36 | 98.75 | 97.98 | 98.31 | **98.78** | 98.06 | **97.69** | 98.36 | 98.12 | **97.92** | **98.65** | 98.55 | 98.63 |
| SCNN_UNet_ Attention _PL** | 98.35 | **98.77** | 97.98 | 98.30 | 98.70 | 98.17 | 97.57 | 98.56 | **98.19** | 97.74 | 98.61 | 98.51 | **98.64** |

*Results reported in [11].   **Results of the models with the proposed self pre-training.*
*"CE" is short for weighted cross entropy loss, and "PL" is short for PolyLoss.*
*Therefore, "UNet_ConvLSTM_CE**" means UNet_ConvLSTM model with self pre-training and using weighted cross entropy loss in the fine-tuning phase. This naming rule applies to all other models.*

(with ** in their names), they take all the best accuracies in all 12 challenging scenes; SCNN_UNet_ConvLSTM_PL** obtains the best overall precision (0.8444) followed by SCNN_UNet_Attention_PL** (0.8413), and also together with other proposed models, they fill 11 best precisions out of all the 12 challenging scenes except for only scene 9 *blur*.

It is worth noting that the models using the proposed self pre-training deliver slightly worse F1-measures compared to those without pre-training. This is because the models are more strict with the pixels classified as the lane lines which might increase the number of lane pixels that are incorrectly identified as background, i.e., resulting in higher false negatives, thus the recall ratio decreases and the F1-measures get slightly worse (even if there are increases in precisions). From Fig. 7 (B), it is more intuitive to see that the developed models with the proposed pre-training and PolyLoss still show acceptable results better than the baselines.

*D. Ablation Study and Discussion*

**Masking ratio:** Experimental results in the previous study [32] showed that the masking ratio needs to correspond to the mask patch, i.e., "for a small mask patch size of 8, the masking ratio needs to be as high as 80% to perform well", while "for a large masking patch size of 32, the approach can achieve competitive performance in a wide range of masking ratios (10%-70%)". In this study, the patch size is set as 16, i.e., (16×16), and the experimental comparisons were carried out with ratios set as 25%, 50%, and 75%.

Testing on SCNN_UNet_ConvLSTM model, Fig. 8 (a) shows the average normalized reconstruction loss indicated by mean square error (MSE) of the image reconstruction task during the pre-training phase. It is observed that using a smaller masking ratio leads to lower reconstruction loss, which is easy to understand, as a smaller masking ratio means fewer pixels need to be reconstructed.

Fig. 8 (b) shows the lane detection performance on the normal driving scene dataset regarding F1-measure with different masking ratios and TABLE III shows the detailed quantitative results.

It is found that although the result of masking at a 75% ratio achieves the best F1-measure of 0.926 on the normal dataset, it does not perform particularly well on the challenge dataset, where it only achieves an F1-measure of 0.71815 worse than that of masking at a 50% ratio (F1-measure at 0.7327).

Furthermore, referring to the results of the pre-training phase, it is clear that masking at a 50% ratio delivers balanced results during both the pre-training phase and fine-tuning testing phases. It is more reasonable to adopt the balanced setting to verify the proposed lane detection pipeline and method, and thus, 50% was chosen as the masking ratio for all testing models.

**Loss function:** Earlier mentioned in this paper, two loss functions (i.e., weighted cross entropy loss and PolyLoss) were tested in the experiments under the proposed pipeline in the fine-tuning segmentation phase. The quantitative comparison results are shown in Table II, and the qualitative results are intuitively demonstrated with visualizations in Fig. 7.

As shown in Table II (a), testing on tvtLANE test set#1 (normal scene), for both SCNN_UNet_ConvLSTM and UNet_ConvLSTM based models, the overall performance of using PolyLoss outperforms that of weighted cross entropy loss. To be specific, compared with UNet_ConvLSTM_CE**, the UNet_ConvLSTM_PL** model obtains an increase of 0.15% in accuracy; a significant increase of 0.039, i.e., around 4.4% improvement, in precision; while a bit decrease in recall; and, overall, a better F1-Measure of 0.915 over 0.910. SCNN_UNet_ConvLSTM_PL** gets the same superiority patterns over SCNN_UNet_ConvLSTM_CE**, and SCNN_UNet_ConvLSTM_PL** obtains the second-best F1-measure (0.922), the second-best precision (0.929), and the best accuracy (98.38%), among all tested models. SCNN_UNet_Attention_PL** slightly beats SCNN_UNet_ConvLSTM_PL** in F1-Mesure (0.924) and precision (0.937). The superiority of the customized PolyLoss over weighted cross entropy loss can be explained by that the PolyLoss function is designed as a linear combination of polynomial functions so that the importance of polynomial bases can be adjusted according to the imbalanced dataset and regarding the segmentation task. With the fine-tuned hyperparameters $\alpha$, $\gamma$, $\varepsilon$ in (9), the customized PolyLoss is perfectly adjusted to the dedicated lane detection task.

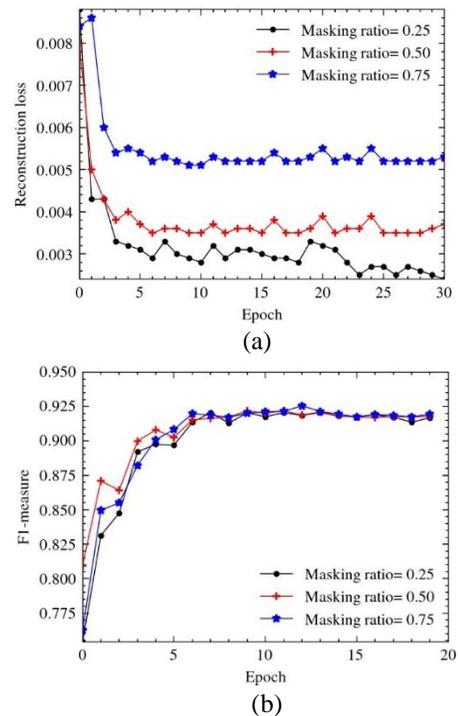

**Fig 8.** Model performance comparison with different masking ratio settings: reconstruction loss in pre-training phase (a), and the F1-measure testing on tvtLANE test set #1 (b)

The model using PolyLoss also performs better than the ones using weighted cross entropy loss in almost all challenging scenes regarding accuracy and precision. In particular, testing on the challenging driving scenes dataset, UNet_ConvLSTM_PL** gets the highest overall accuracy at 98.38%, while SCNN_UNet_ConvLSTM_PL** obtains the best overall precision at 0.8444.

**Training time and model complexity**: In addition to the improvement regarding the evaluation metrics, the proposed

TABLE III
MODEL PERFORMANCE WITH DIFFERENT MASKING RATIOS
(a) tvtLANE TEST SET #1 (NORMAL)

| Mask Ratio | Test_Acc (%) | Precision | Recall | F1-Measure |
|---|---|---|---|---|
| 25% | 98.36 | 0.927 | 0.915 | 0.921 |
| 50% | 98.20 | 0.891 | **0.952** | 0.921 |
| 75% | **98.40** | **0.933** | 0.918 | **0.926** |

(b) tvtLANE TEST SET #2 (CHALLENGING)

| Challenging Scenes | Precision | | | F1-Measure | | | Accuracy (%) | | |
|---|---|---|---|---|---|---|---|---|---|
| Mask ratio | 25% | 50% | 75% | 25% | 50% | 75% | 25% | 50% | 75% |
| overall | 0.8248 | 0.8001 | 0.8348 | 0.7196 | 0.7327 | 0.7162 | 98.31 | 98.03 | 98.36 |
| 1-crve&occlude | 0.8083 | 0.8754 | 0.9433 | 0.8238 | 0.8937 | 0.9260 | 98.58 | 98.33 | 98.83 |
| 2-shadow-bright | 0.8881 | 0.8672 | 0.9028 | 0.7953 | 0.869 | 0.8777 | 98.01 | 97.64 | 98.09 |
| 3-bright | 0.8611 | 0.8519 | 0.8786 | 0.7944 | 0.8426 | 0.8111 | 98.30 | 98.05 | 98.34 |
| 4-occlude | 0.8480 | 0.7763 | 0.8615 | 0.7703 | 0.7656 | 0.7438 | 98.80 | 98.45 | 98.83 |
| 5-curve | 0.9327 | 0.8664 | 0.9187 | 0.7660 | 0.8352 | 0.8840 | 98.14 | 97.69 | 98.14 |
| 6-dirty&occlude | 0.7052 | 0.5523 | 0.4813 | 0.3595 | 0.2493 | 0.2655 | 97.55 | 97.42 | 97.44 |
| 7-urban | 0.5090 | 0.5261 | 0.5565 | 0.4939 | 0.5751 | 0.5150 | 98.46 | 97.95 | 98.52 |
| 8-blur&curve | 0.7915 | 0.7396 | 0.7823 | 0.7933 | 0.7756 | 0.7426 | 98.01 | 97.54 | 98.08 |
| 9-blur | 0.9473 | 0.8865 | 0.9462 | 0.7396 | 0.7122 | 0.7437 | 97.74 | 97.57 | 97.76 |
| 10-shadow&dark | 0.9553 | 0.8974 | 0.9331 | 0.7942 | 0.7661 | 0.7180 | 98.71 | 98.38 | 98.65 |
| 11-tunnel | 0.8427 | 0.8115 | 0.8956 | 0.7217 | 0.6989 | 0.6667 | 98.44 | 98.23 | 98.53 |
| 12-dim&occlude | 0.7588 | 0.9101 | 0.8750 | 0.6173 | 0.542 | 0.5973 | 98.33 | 98.56 | 98.54 |

*All of the test results in TABLE III were tested on the SCNN_UNet_ConvLSTM model*

self pre-training pipeline plus the customized PolyLoss can also reduce the training time with the model convergence speed greatly improved. To be specific, tests revealed for UNet_ConvLSTM based models, UNet_ConvLSTM_PL** converged at the 10$^{th}$ epoch, while UNet_ConvLSTM_CE** converged at the 91$^{st}$ epoch, and UNet_ConvLSTM without the proposed pertaining needed around 100 epochs to converge [1]. Similarly, for SCNN_UNet_ConvLSTM based models, SCNN_UNet_ConvLSTM_PL** converged at the 12$^{th}$ epoch, while SCNN_UNet_ConvLSTM_CE** converged at the 29$^{th}$ epoch, and SCNN_UNet_ConvLSTM without the proposed pre-training needed around 100 epochs to converge.

These results demonstrate that pre-training with masked sequential autoencoders plus fine-tuning with PolyLoss can not only boost the models' overall performance regarding accuracy, precision, and F1-measure, but also speed up model convergence greatly reducing the training time.

Furthermore, from the parameters and MACs illustrated in Table II (a), it is demonstrated that, with the proposed pre-training and customized PolyLoss, the model size and complexity merely change.

In short, the proposed pipeline contributes to the improvement of model efficiency and detection accuracy simultaneously.

## V. CONCLUSION

In this paper, a novel deep learning pipeline integrating self pre-training with masked sequential autoencoders, fine-tuning segmentation with customized PolyLoss, and post-processing with clustering and curve-fitting, is proposed for the vision-based robust lane detection task. With the proposed self pre-training method by reconstructing the randomly masked image frames and the customized PolyLoss for the fine-tuning segmentation phase, the tested three neural network models (i.e., UNet_ConvLSTM, SCNN_UNet_ConvLSTM, and SCNN_UNet_Attention) all delivered significantly better performances in comparison to baselines. Through extensive experiments, the models under the proposed pipeline surpass other state-of-the-art models with the best testing accuracy, precision, and F1-measure on the normal driving dataset (i.e., tvtLANE test set #1) and the best overall accuracy and precision on the 12 challenging driving scenarios (tvtLANE test set #2). Furthermore, without changes in the model size and complexity, under the proposed pipeline, the test models converged faster, especially when adopting the customized PolyLoss in the fine-tuning segmentation phase, while performing better detection results. These findings demonstrate the effectiveness of the proposed lane detection pipeline which upgrades the model training efficiency and detection accuracy simultaneously.

It is witnessed that when testing with some brand new challenging samples, i.e., no similar samples are covered in the training phase, the model might be defeated with a low F1-measure. In practice, lane detection models trained on datasets from one certain country might not work well when testing on datasets with different lane structures from another country. To tackle this problem and further enhance the model's robustness, for future studies, it is suggested to investigate domain generalization and adaption methods to transfer the knowledge and patterns learned from available datasets to unseen domains and fields with brand new data.

ACKNOWLEDGMENT

The authors would like to thank Dr. Haneen Farah for her valuable comments and suggestions on this work.

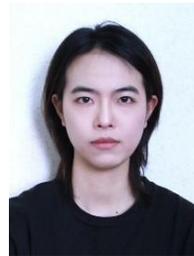

**Ruohan Li** received the B. Eng. degree and M.Eng. degree in traffic and transportation from Lanzhou Jiaotong University, Lanzhou, China, in 2020 and 2023. She is currently pursuing the Ph.D. degree and working as a graduate research assistant in the Department of Civil and Environmental Engineering, Villanova University. Her research interests include deep learning, automated driving, and transportation big data.

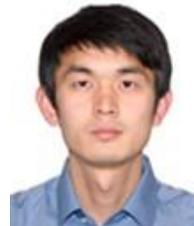

**Yongqi Dong** received the B.S. degree in telecommunication engineering from Beijing Jiaotong University, Beijing, China, in 2014 and the M.S. degree in control science and engineering from Tsinghua University, Beijing, China, in 2017. He is currently working as a Ph.D. researcher with the Department of Transport and Planning, Delft University of Technology. He is also a visiting scholar at the Department of Mechanical Engineering, University of California, Berkeley. His research interests include deep learning, transportation big data, automated driving, and traffic safety. He seeks to employ artificial intelligence and interdisciplinary research as tools to shape a better world.